\documentclass[11pt,a4paper]{article}
\usepackage[hyperref]{eacl2021}
\usepackage{times}
\usepackage{latexsym}

\usepackage{microtype}
\usepackage{amsmath}
\usepackage{amssymb}
\usepackage{bbm}
\usepackage{graphicx}
\usepackage{multicol}
\usepackage{tabularx}
\usepackage{booktabs}
\usepackage{algorithmic}
\usepackage{algorithm}
\usepackage{cleveref}
\usepackage{multirow}

\usepackage{multirow, tabu}

\crefformat{section}{\S#2#1#3}
\crefformat{subsection}{\S#2#1#3}
\crefformat{subsubsection}{\S#2#1#3}
\DeclareMathOperator*{\argmax}{arg\,max}

\definecolor{darkgreen}{RGB}{84,174,50}

\aclfinalcopy 
\def\aclpaperid{325} 

\title{Neural Data-to-Text Generation with LM-based Text Augmentation}

\author{ 
  \dag Ernie Chang, $\odot$ Xiaoyu Shen\thanks{ $\>$ Work done prior to joining Amazon.}
, \dag Dawei Zhu, \dag Vera Demberg, $\otimes$ Hui Su \\
  \dag Dept. of Language Science and Technology, Saarland University \\
    {\tt \{cychang,xiaoyu\}@coli.uni-saarland.de}
  \\ $\odot$ Amazon Alexa AI, Berlin \\ $\otimes$ Pattern Recognition Center, Wechat AI, Tencent Inc, China
  \\ 
}

\date{}

\begin{document}
\maketitle
\begin{abstract}
For many new application domains for data-to-text generation, the main obstacle in training neural models consists of a lack of training data. While usually large numbers of instances are available on the data side, often only very few text samples are available. To address this problem, we here propose a novel few-shot approach for this setting. Our approach automatically augments the data available for training by (i) generating new text samples based on replacing specific values by alternative ones from the same category, (ii) generating new text samples based on GPT-2, and (iii) proposing an automatic method for pairing the new text samples with data samples. As the text augmentation can introduce noise to the training data, we use \textit{cycle consistency} as an objective, in order to make sure that a given data sample can be correctly reconstructed after having been formulated as text (and that text samples can be reconstructed from data).

On both the E2E and WebNLG benchmarks, we show that this weakly supervised training paradigm is able to outperform fully supervised seq2seq models with less than $10\%$ annotations. 
By utilizing all annotated data, our model can boost the performance of a standard seq2seq model by over 5 BLEU points, establishing a new state-of-the-art on both datasets.
\end{abstract}

\section{Introduction}
\begin{figure}
  \centering
\includegraphics[width=1.\columnwidth]{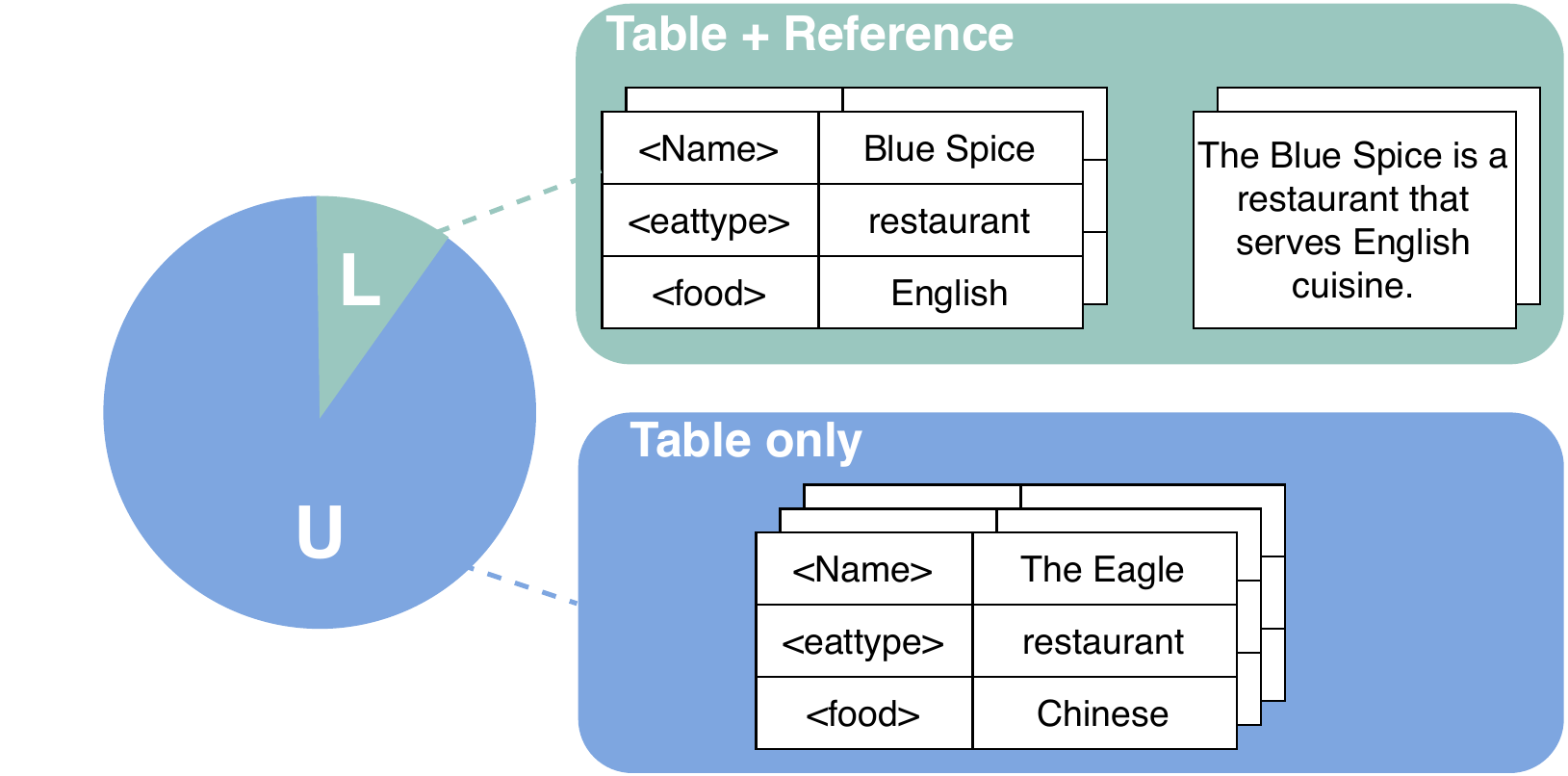}
\caption{ \small
\textbf{Few-shot scenario}: The model is expected to learn data-to-text generation with few labeled instances (i.e. table-text pairs). The example is taken from the E2E dataset.}
\label{fig:sample}
\end{figure}

Neural data-to-text generation has been the subject of much recent research. 
The task aims at transforming \emph{source-side} structured data into \emph{target-side} natural language text~\cite{reiter2000building,barzilay-lapata-2005-modeling}. 
While neural end-to-end systems afford the advantage of easy adaptability~\cite{lebret2016neural,wiseman2017challenges}, huge amounts of data-text pairs are still necessary to perform on par with their rule-based counterparts \cite{van2018automated}. 
This makes using neural systems less appealing: oftentimes, in-domain text samples are not readily available, and there is a high cost to collecting in-domain texts which fit the data samples, and annotating these texts with the data labels -- the cost for collecting this data might hence even outweigh the efforts of designing a 
rule-based system~\cite{gkatzia2016content}.
The goal of this work is to improve the performance of neural data-to-text models in scenarios where only very few text samples exist (we assume that these text samples are paired with corresponding data samples). 
We aim to answer how we can make the most of the scarce annotations, together with large amounts of unlabelled data, in order to push the limit of the neural data-to-text models. Figure~\ref{fig:sample} illustrates the scenario.

To address the limited-data challenge, we propose a simple yet effective way of augmenting the text side with the pretrained language model (LM) GPT-2~\cite{gpt2}. Unlike other text augmentation work employed in data-to-text generation systems~\cite{freitag2018unsupervised,agarwal2018char2char}, our proposal assumes little to no domain-dependent heuristics.
It consists of two steps: 
(1) information augmentation by slot-value replacement and 
(2) LM augmentation by GPT-2 generation. 

Once we have augmented the set of text samples, we are essentially in a similar setting as previously proposed semi-supervised approaches to data-to-text generation \citet{schmitt2019unsupervised,qader2019semi,su2020towards}, which assume the presence of vast amounts of \textit{unpaired} data and text instances. These approaches exploit a \textit{cycle consistency} objective in order to learn a pairing for the data samples. The cycle consistency objective tries to make sure that data samples can be reconstructed correctly from their textual formulations, and similarly that texts can be reconstructed after having been parsed into a data representation. 

As the automatically generated text samples from GPT-2 might be very noisy and not pair well with data samples, we align each augmented text sample with its most similar unlabeled data sample, as defined in their encoded vector space. This idea is inspired by recent work on representation matching in MT~\cite{artetxe2019margin,ruiter2019self}. To ensure good quality of the training data, only pairs above a certain similarity threshold $\epsilon$ are retained as pseudo pairs for training. 
The quality of the pseudo pairs will gradually improve as the encoder improves in the training process. In return, the learning of the encoder will also be facilitated with the improved quality of pseudo pairs as a virtuous cycle.

On two data-to-text benchmarks E2E~\cite{novikova2017e2e} and WebNLG~\cite{gardent2017webnlg}, we show that our LM-augmented weakly supervised model succeeds on outperforming fully supervised seq2seq model, though utilizing less than $10\%$ of the data annotations. 
It even outperforms previous work which \emph{additionally has access to all unpaired text samples}. 
When trained with full data annotations, it is able to boost the model performance by up to 5 BLEU points, establishing a new state-of-the-art on both datasets. 

In summary, this work makes the following contributions:
\begin{enumerate}
    \item We study the few-shot data-to-text scenario where, unlike previous works, no further target-side text is available.
    \item We present an effective way of automatically augmenting target text by resorting to the pretrained LM GPT-2.
    \item We propose utilizing the augmented text by a combination of cycle consistency and representation matching. The resulting model outperforms standard seq2seq model with less than 10\% data annotations.
    \item The proposed model is shown to be complementary with current seq2seq pretraining techniques, and can offer orthogonal improvements when combining both.
\end{enumerate}

\section{Related Work}
Building neural data-to-text systems with few paired samples (but a large set of unpaired samples) has been a hot research topic recently. 
Most works adopt the idea of cycle consistency~\cite{zhu2017unpaired}, which has been used in many text generation tasks like machine translation~\cite{artetxe2017unsupervised,lample2017unsupervised} and style transfer~\cite{prabhumoye2018style,subramanian2018multiple}. 
\citet{schmitt2019unsupervised,qader2019semi,su2020towards,chang2020dart,chang2021jointly,chang2021does} applied this idea to the task of data-to-text generation and reported promising results. 
\citet{ma2019key} separate the generation process into few-shot content selection and surface realization components and learn them separately. 
Nonetheless, all of these approaches assume the existence of \emph{huge quantity of unpaired text samples}, which, as we mentioned, is an unrealistic assumption for the task of data-to-text generation.
 \citet{freitag2018unsupervised} proposes to reconstruct usable sequences re-written from data with rules for unsupervised data-to-text generation. 
Unfortunately, designing these rules require efforts similar to building a template-based system. 
\cite{budzianowski2019hello,chen2019few,peng2020few} tackle the few-shot challenge by finetuning a pretrained LM to incorporate prior knowledge from general-domain text or data-text pairs. 
We show that our technique is complementary with them and can offer orthogonal improvements when combining both.

\section{Problem Formulation}
\label{sec:length}
We represent the data samples as \textit{D} and the text samples as \textit{T}. In our work, we do not restrict the format of the data. Each $d \in D$ can be a set of key-value pairs, as in Figure~\ref{fig:sample}, or in form of RDF triples as in \citet{gardent2017webnlg}. Each text $t \in T$ consists of a sequence of words.
In few-shot settings, we are assumed to have (1) $k$ labeled pairs $(D_L, T_L)$ and (2) large quantities of unlabeled data $D_U$ where $|D_U| \gg k > 0$\footnote{We force $k>0$ as we believe a reasonable generation system needs a least a few demonstrations of the annotation.}. 
This, we believe, is a more realistic setting as unlabeled data are usually abundant and also can be easily fabricated from predefined schemata. Notably, we assume no access to outside resources containing in-domain text. The $k$ annotations are all we know about the text side.

\section{Approach}
%

In this section, we first explain our proposed new method for text sample augmentation, and then discuss methods to remove noise and automatically align the data by elaborating on the ideas of cycle consistency and representation matching.  Finally, we summarize the approach and present the detailed algorithm.

\begin{figure*}[t!]
  \centering
\includegraphics[width=0.7\textwidth]{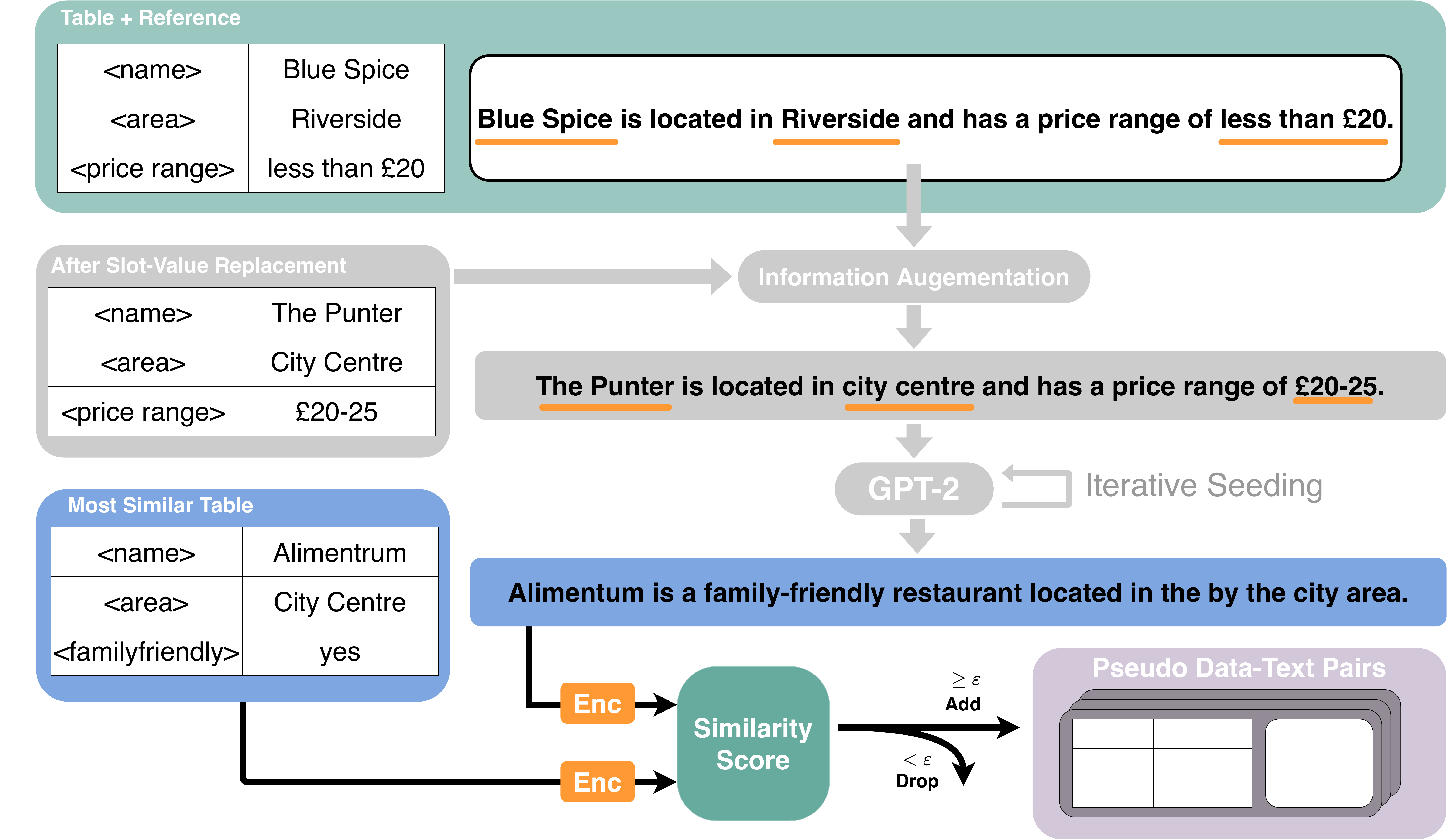}
\caption{\label{fig:augment}\small Depiction of \emph{text augmentation} and \emph{representation matching}. Each text sample first goes through information augmentation by slot-value replacement, then passed to GPT-2 with iterative conditional generation. The augmented text samples are paired with the most similar data from the corpus with a threshold cutoff. }
\label{in_text_sample}
\end{figure*}

\subsection{Text Augmentation}
\label{sec: tt-aug}
To mitigate the paucity of the set of text samples $T$, we propose a pipeline approach to augment the text samples by (1) information augmentation and (2) LM augmentation.

\subsubsection{Information Augmentation}
\label{sec: info-aug}
We generate additional text samples by performing slot-value replacements. As many data values are exactly copied to the text samples, these copied information can be easily detected and replaced with other values (for the same slot type) to enrich the information space of the text samples. 
This can be considered as a simplified version of traditional methods of template mining where key words are extracted to construct templates ~\cite{kondadadi2013statistical,oya2014template}. An example is shown in Figure~\ref{fig:augment}. Each text sample is augmented with 10 more distinct text samples or with all possible values being replaced.

The slot-value replacement is efficient to implement. However, it can only detect identical values and augment text with the same combinatorial patterns as the few-shot annotations. To enrich the linguistic realizations of text sentences and enable new combinations of information, we further propose a LM augmentation approach using GPT-2.

\subsubsection{LM Augmentation}
\label{sec: lm-aug}
 GPT-2 \cite{gpt2} is a language model pretrained on the collected WebText. It has demonstrated remarkable zero-shot multitask adaptability by simply feeding the input of each task into the LM and continuing to generate words. People have also also shown that GPT-2 is able to improve classification tasks via in-domain text augmentation~\cite{papanikolaou2020dare,sun2020lamal}. We use a similar technique by first fine-tuning GPT-2 in the few-shot annotations~\cite{Wolf2019HuggingFacesTS}, and then applying it to produce synthetic text through an iterative conditional generation process: With initial seeds being samples of $T_{L}$ plus new samples  from information augmentation, the LM iteratively conditions on the previous output sentence to generate in-domain text\footnote{We adopt the Top-$k$ random sampling setting with $k=2$ to encourage diversity and reduce repetition~\cite{gpt2}}. Each synthetic sentence is pruned if it (1) is shorter than 5 words or (2) contains only special tokens. The iterative generation is terminated when all tokens in the initial seeds are covered 
 or if the maximum of 100 runs is reached. All the unpruned synthetic text samples are added into the space of $T$ to 
benefit the learning direction of $t \rightarrow d' \rightarrow t$ and $\Tilde{t}\rightarrow t$.
Figure~\ref{in_text_sample} depicts the generation process of GPT-2.

In practice, obtaining clean in-domain text requires extreme efforts of designing heuristic rules. Nonetheless, the synthetic text from GPT-2 makes decent sense and can already provide useful signals to drive the learning process. 

\subsection{Cycle Consistency}
\label{sec: cycle}
The core idea of encouraging cycle consistency is that starting from one sample in a domain, the model first maps it into the other domain, then maps it back~\cite{he2016dual}. The resulting sample should be identical to the original sample. Specifically, let $p_\theta(t|d)$ be the probability distribution to map a data sample $d$ to its corresponding text $t$, and $p_\phi(d|t)$ be the probability distribution to map text back to data. Starting from a data sample $d \in D$, its objective is:
\begin{equation}
\label{eq: dtd}
\begin{split}
    \max_{\phi}      \mathbb{E}_{d \sim p(D)}\log p_\phi(d|t');\;
    t' \sim p_\theta(t|d)
\end{split}
\end{equation}
which basically ensures the consistency in the direction of $d \rightarrow t' \rightarrow d$. Note that only $p_\phi$ is updated in this direction and $p_\theta$ serves only as as an auxiliary function to provide pseudo samples $t'$ from $d$. Though it is also possible to update $\theta$ at the same time through tricks like Gumbel-softmax~\cite{jang2016categorical} or REINFORCE~\cite{williams1992simple}, we find it did not lead to better performance, yet complicated the training. Similar observations have been made in \citet{lample2018phrase,he2020probabilistic,garcia2020multilingual}.

Similarly, starting from a text $t \in T$, the objective is to ensure the consistency in the direction of $t \rightarrow d' \rightarrow t$:\footnote{In the MT community, the equivalent step is usually called \emph{back translation}~\cite{sennrich2016improving,lample2018phrase}.}
\begin{equation}
\label{eq: tdt}
\begin{split}
    \max_{\theta}      \mathbb{E}_{t \sim p(T)}\log p_\theta(t|d');\;d' \sim p_\phi(d|t)
\end{split}
\end{equation}
Finally, we further add two denoising autoencoding objectives on both the data and text sides:
\begin{equation}
\label{eq: ae}
    \max_{\theta,\phi} \mathbb{E}_{d \sim p(D), t \sim p(T)} \log p_\phi(d|\Tilde{d})p_\theta(t|\Tilde{t})
\end{equation}
where $\Tilde{d}$ and $\Tilde{t}$ are the corrupted versions of $d$ and $t$. We use the same noise function as in \citet{lample2018phrase} which randomly permutes and pads a portion of the input. This can encourage the encoder to learn meaningful latent representations by reconstructing the input itself~\cite{currey2017copied,lample2018phrase}. 

\begin{figure}[t!]
  \centering
\includegraphics[width=\columnwidth]{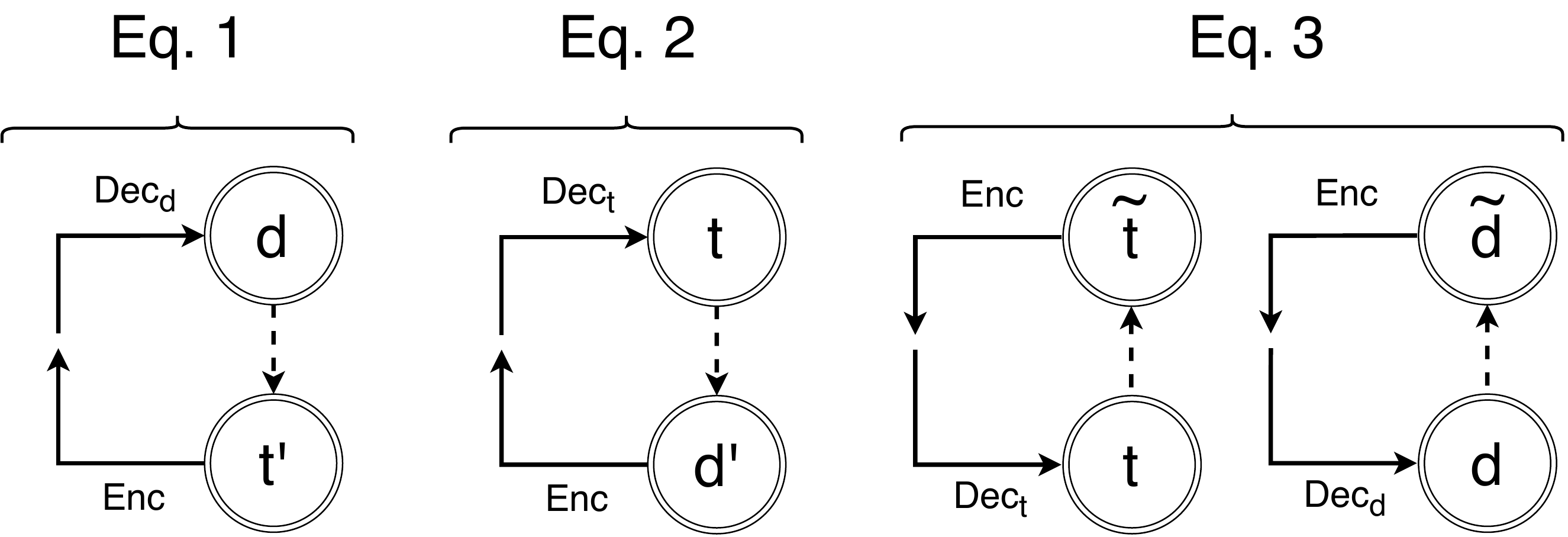}
\caption{\small Four directions of cycle consistency. Gradients are backpropagated only through solid lines.}
\label{fig:directions}
\end{figure}

Figure~\ref{fig:directions} illustrates all the four directions of the cycle consistency objective.
%

We use one shared encoder $Enc$ for both the data and text sides. Each data sample is flattened into a sequence 
by making a list of slot value pairs 
and fed into the same encoder. Using the same encoder for both types of input gives the model an inductive bias to project similar data/text into surrounding latent space.

We will show later that encoder sharing is essential for a good performance under the few-shot scenario. From the shared encoded space, two separate decoders $Dec_d$ and $Dec_t$ are used to decode $d$ and $t$ respectively\footnote{The shared encoding has also been shown effective in other tasks like machine translation~\cite{lample2018phrase} and image transition~\cite{zhu2017unpaired}. We further tried sharing the decoder as in \citet{johnson2017google} but find no improvement (see Table~\ref{tb:dir}).}.

\subsection{Representation Matching}
    \label{sec: rm}
Apart from training under the cycle consistency, we further consider matching each synthetic text with its most similar data sample and treating them as supplementary training pairs. Compared with the pseudo $d'$ obtained from back translation
(Eq.~\ref{eq: tdt}), the matched data samples are extracted from the existing corpus $D_U$ and thereby are guaranteed to be clean. This can provide a much more stable training signal especially at the initial training stage\footnote{In theory, as we can fabricate arbitrary possible data samples from the predefined schema and add to the corpus, we can always find one matched data for a text samples.}.
Previous work has used representation matching to automatically extract pseudo training pairs for machine translation~\cite{artetxe2019margin,ruiter2019self}. \citet{baziotis2019seq3,chu2019meansum} also demonstrate that the representation similarity between input-output pairs can serve as a useful regularization for unsupervised text summarization. We adopt a similar idea to create pseudo pairs based on their cosine similarity in the representation space. To summarize, the process of representation matching can be described as:
\begin{equation}
\label{eq: rm}
\begin{split}
    \max_{\theta,\phi} \mathbb{E}_{t \sim p(T')} \mathbbm{1}_{cos(d^{\ast},t) > \varepsilon}(&\log p_\theta(t|d^{\ast}) \\ + &\log p_\phi (d^{\ast}|t));\\
    d^{\ast} = \argmax_{d \in D}\; &\text{cos}(d,t)
\end{split}
\end{equation}
where $T'$ is augmented text from the LM and $\mathbbm{1}$ is the indicator function. We also perform mean pooling over the encoded representations before matching them. $\varepsilon$ is a threshold. Pseudo pairs with a cosine similarity less than $\varepsilon$ will be discarded. Ideally, as the encoder improves, the pseudo pairs created by representation matching will make more sense, which can in turn benefit the training of the encoder. 


\subsection{Summary}
\label{sec: summary}
Apart from the above unsupervised objective, on the few annotated data-text pairs, we can impose the supervised objective:
\begin{equation}
\label{eq: sup}
    \max_{\theta,\phi} \mathbb{E}_{d,t \sim p(D_L,T_L)} \log p_\theta(t|d) + \log p_\phi (d|t)
\end{equation}
where $(D_L,T_L)$ contains the $k$ data annotations. Putting all together, we summarize it in Algorithm 1. In the training stage, we optimize the objectives of cycle consistency, representation matching and supervised learning sequentially to maintain a constant ratio of signals from all sides.

\begin{algorithm}[tb]
   \caption{Few-shot Data-to-text Framework}
   \label{alg}
\begin{algorithmic}[1]
   \STATE {\bfseries Input:} $D_U, (D_L, T_L)$
   \STATE Create $(D_a, T_a)$ by information augmentation ;
   \STATE $(D_L, T_L) \leftarrow (D_L, T_L) \cup (D_a, T_a)$;
   \STATE Create $T'$ by LM augmentation ;
   \STATE $T \leftarrow T_L \cup T'$;
   \REPEAT
   \STATE Sample batch data from $(D_L, T)$ ;
   \STATE \textbf{Cycle consistency:}
   \STATE Optimize by Eq.~\ref{eq: dtd} + Eq.~\ref{eq: tdt} + Eq.~\ref{eq: ae};
   \STATE \textbf{Representation Matching:}
   \STATE Optimize by Eq.~\ref{eq: rm};
   \STATE \textbf{Supervised Training:}
   \STATE Optimize by Eq.~\ref{eq: sup};
   \UNTIL{convergence}
\end{algorithmic}
\end{algorithm}

\section{Experiment Setting}
\label{sec:exp}
\paragraph{Data}
We conduct experiments on the E2E~\cite{novikova2017e2e} and WebNLG~\cite{colin-etal-2016-webnlg} datasets. E2E is a crowd-sourced dataset containing 50k instances in the restaurant domain. The inputs are dialogue acts consisting of three to eight slot-value pairs. WebNLG contains 25k instances describing entities belonging to fifteen distinct DBpedia categories. The inputs are up to seven RDF triples of the form \emph{(subject, relation, object)}. 

\paragraph{Configuration}
The model is implemented based on fairseq~\cite{ott2019fairseq}.
We use $600$-dimensional token embedding and Adam optimizer with initial learning rate at $0.0002$. 
Batch size is kept at $48$ with a dropout rate at $0.3$. 
We employ beam search with size $3$ for decoding and select
models based on BLEU-4 scores on the development set. The score is averaged over 10 random initialization runs. 
In this work, the seq2seq models are built upon the long short-term memory (LSTM)~\cite{hochreiter1997long}. For LSTM cells, both the encoder and decoder have $3$ layers, amounting to 18M parameters for the seq2seq model (600-dimension and 1024 hidden units). Maximum sequence length is set as 100 for E2E and 200 for WebNLG (SPM-based). All encoder parameters are shared between data and text samples. 
All models were trained on 1 Nvidia V100 GPUs (32GB and CUDA Version 10.2) for 4k steps. 
The total batch size is around $48$K tokens per GPU and we use the Adam optimizer ($\epsilon=1\mathrm{e}{-6}$, $\beta_2 = 0.98$) along with linear learning rate decay scheduling. 
The total number of updates is set to $8000$ for all training and models are selected based on optimal validation BLEU4. At decoding time, sentences are generated using greedy decoding. 


\section{Results and Analysis}

\begin{table*}
  \small
  \centering
  \resizebox{0.9\textwidth}{!}{
  \begin{tabular}{lcccccccc}
    \toprule
    \multirow{2}{*}{Model} & \multicolumn{4}{c}{E2E - 10\%} & \multicolumn{4}{c}{E2E - 100\%} \\
    \cmidrule(r){2-5}
    \cmidrule(r){6-9}
     & BLEU & NIST & METEOR & ROUGE-L & BLEU & NIST & METEOR & ROUGE-L \\
    \cmidrule(r){1-1}
    \cmidrule(r){2-5}
    \cmidrule(r){6-9}
    SLUG & - & - & - & - & 66.19 & 8.61 & 44.54 & 67.72\\
    Seq2seq & 53.38 & 6.10 & 38.10 & 60.53 
    & 63.32 & 6.81 & 41.25 & 62.91 \\
    \citet{qader2019semi} & 58.10& 6.24 & 41.32 & 62.84 
    & 64.20 & 7.14 & 44.68 & 65.31 \\
    \citet{chen2019few} & 59.10 & 7.49 & 40.25 & 63.23
    & 63.72 & 7.76 & 40.25 & 66.23\\
    Proposed (LSTM) & 64.24  & 7.71 & 43.53 & 66.81 
    & 68.88  & 8.89 & 48.53 & 72.12\\
    \midrule
    \midrule& \multicolumn{4}{c}{WebNLG - 10\%} & \multicolumn{4}{c}{WebNLG - 100\%} \\
    \cmidrule(r){2-5}
    \cmidrule(r){6-9}
    Model & BLEU & NIST & METEOR & ROUGE-L & BLEU & NIST & METEOR & ROUGE-L \\
    \cmidrule(r){1-1}
    \cmidrule(r){2-5}
    \cmidrule(r){6-9}
    Melbourne & - & - & - & - 
    & 44.93 & 8.98 & 36.58 & 60.40\\
    Seq2seq & 36.54 & 7.3 & 35 & 54.61
    & 44.60 & 8.49 & 38.23 & 59.67\\
    \citet{qader2019semi} & 38.66 & 7.81 & 34.1 & 56.95
    & 47.19 & 8.71 & 37.90 & 58.61\\
    \citet{chen2019few} & 39.40 & 7.84 & 37.25 & 56.23
    & 46.15 & 8.52 & 39.1 & 58.5\\
    Proposed (LSTM) & 43.75  & 8.29 & 33.58 & 58.49
    & 50.26  & 8.86 & 40.71 & 61.29\\
    \bottomrule
  \end{tabular}}  
\caption{ \small Performance on E2E and WebNLG with 10\% and 100\% data. \citet{qader2019semi} utilizes all ground-truth unpaired text samples while our proposed model only gets access to the few-shot data annotations.}
\label{tb:comparison}
\end{table*}
In this section, we present experiment results and analysis. We first compare our model with other baselines on both datasets, then perform a set of ablation studies on the E2E dataset to see the effects of each component. Finally, we analyze how  text augmentation helps improves the model, include example outputs and show the human evaluation results in the end.
\paragraph{Comparison with Other Models}
In Table~\ref{tb:comparison}, we compare our model with (1) seq2seq baseline, (2) cycle consistency model as in \citet{qader2019semi}\footnote{The author did not open-source their code. We reproduced their model based on our implementation. The results on 10k annotations matches their reports in the paper.} and (3) finetuned GPT-2 model as in \citet{chen2019few}\footnote{https://github.com/czyssrs/Few-Shot-NLG}. 
For all models, we try running with 10\% and 100\% annotations to see how they perform under different data sizes. Our model is implemented both with LSTM encoder-decoders, same as the seq2seq baseline for a fair comparison.
Note that \citet{qader2019semi} further \emph{utilized all the ground-truth unpaired text samples}, while the other models run only on the few-shot annotations.
We also include the results of \textbf{SLUG}~\cite{juraska2018deep} and \textbf{MELBOURNE}~\cite{gardent2017webnlg}, the overall winner on automatic metrics in the E2E and WebNLG challenge respectively(both seq2seq-based). SLUG uses a heuristic slot aligner based on a set of handcrafted rules and combines a complex pipeline of data augmentation, selection, model ensemble and reranker.

The results show that our proposed model significantly improves over the baseline on both the few-shot and fully supervised setting. The improvement is more evident when only 10\% annotations are available, with a leap of 11 and 7 BLEU scores on E2E and WebNLG respectively. It also outperforms systems relying on task-dependent heuristics. 
In comparison, \citet{qader2019semi}, though with access to all text samples at all percentages, still underperforms our model with tangible margin. On the fully supervised setting, it brings little to no difference compared with the seq2seq baseline as no more extra data is incorporated in the training process. As such, we also observe that the text augmentation from finetuned GPT-2 model helps the proposed model on the few-shot setting, but its advantage also vanishes when all data annotations are available.

\begin{figure*}[ht]
  \centering
\includegraphics[width=0.9\textwidth]{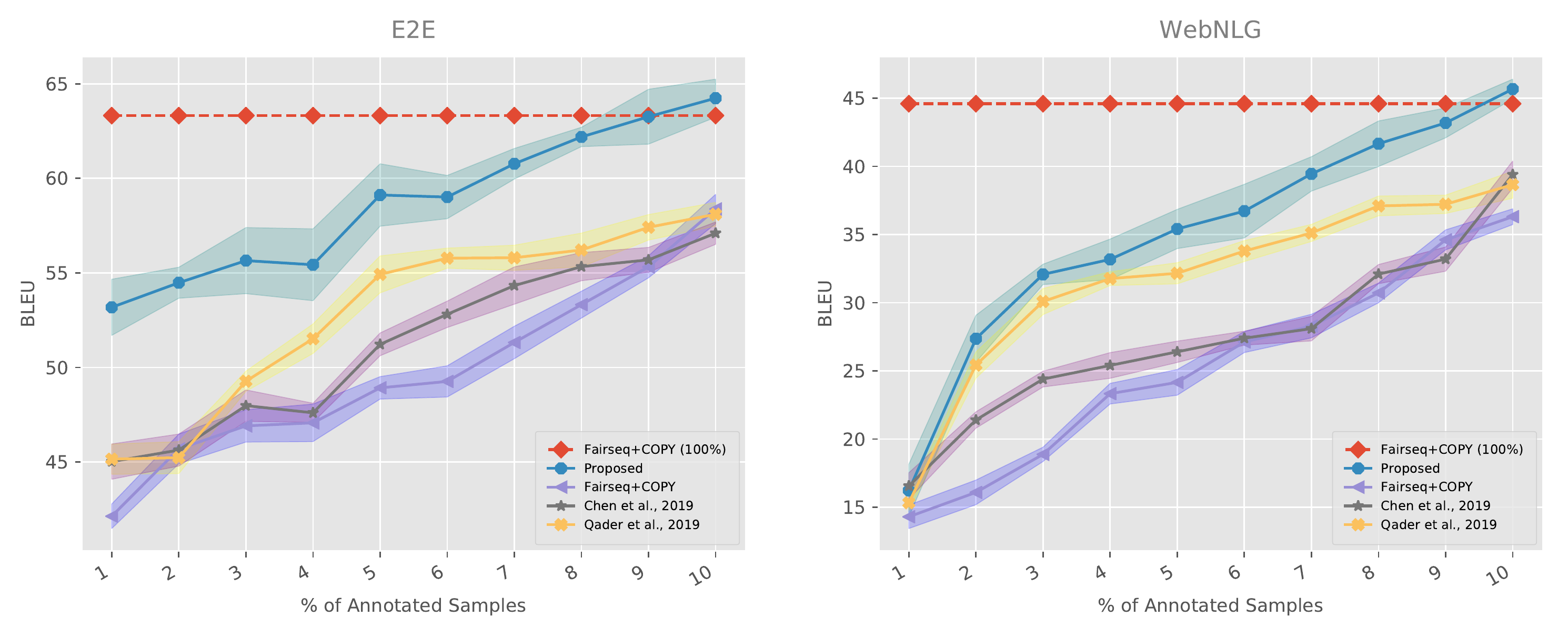}
\caption{\small Model performance with varying number of data annotations. Our model with 10\% annotations outperforms the seq2seq model trained on $100\%$ pairs (dotted line) on both datasets. Shades are the sample standard deviation based on 10 runs of different model initializations. 
}
\label{fig:few}
\end{figure*}

In Figure~\ref{fig:few}, we draw the model performance with varying number of data annotations. All models are trained from scratch with 10 different random initializations and the standard deviation of the BLEU-4 score is visualized. We can see our model (LSTM-based), though with a relatively larger standard deviation due to the uncertainty of text augmentation sampling, still consistently outperforms other baselines significantly and even surpasses the fully supervised seq2seq model with less than 10\% of data annotations.

\begin{table}[t!]
\centering
\resizebox{0.95\columnwidth}{!}{
\begin{tabular}{l|c|c|c|c}

\small\textbf{Model/Share}  & \textbf{\small None}  & \textbf{\small Enc}   & \textbf{\small Dec} & \textbf{\small Both} \\ \hline
 \small \textbf{Supervised}                     & 53.20  & -    & -        & -     \\ 
\textbf{+ $t \rightarrow d' \rightarrow t$}     & 53.19  & 53.28    & 53.17        & \textbf{53.29}    \\ 
 \textbf{+ $d \rightarrow t' \rightarrow d$}    & 53.15  & \textbf{56.12}    & 53.49        & 56.07    \\ 
  + $t \rightarrow t$                   & 53.74  & \textbf{56.39}    & 55.29        & 55.73      \\ 
 + $d \rightarrow d$                    & 53.37  & \textbf{56.44}    & 56.09        & 56.11    \\
 + Noise                                        & 54.13  & \textbf{57.37}    & 56.59        & 57.04  \\
 \hline
\end{tabular}
}
\caption{\label{tb:dir}\small Ablation study for cycle consistency (10\% annotations). BLEU-4 score is reported. Each line adds one condition on top of the previous one. \textbf{Supervised} is a supervised seq2seq baseline.}
\end{table}

\paragraph{Ablation Study on Cycle Consistency}
In Table~\ref{tb:dir}, we study how the four directions, input noise and parameter sharing affect the performance of cycle-consistency. The experiments are conducted with 10\% annotations and \emph{no further unpaired text samples are available}.

As can be observed, adding the training direction $t \rightarrow d' \rightarrow t$ (i.e. back translation) has little effects on top of the supervised seq2seq baseline. This is expected since back translation is naturally designed to incorporate additional unpaired text samples. When run only on the few-shot annotations, its power is very limited. The backward direction $d \rightarrow t' \rightarrow d$ is surprisingly useful when the encoder is shared between the data and text. Though this direction will not affect the text decoder at all, the improvement suggests the model can benefit a lot by simply structuring its encoded space and mapping aligned data-text pairs to similar vector space. The autoencoding directions brings a little improvement. When combined with input noise, the performance further increases. This is similar to previous findings that denoising autoencoding is more helpful in inducing meaningful latent space~\cite{lample2018phrase} in comparison to simply learning to copy the original input. 

The results also suggest encoder sharing is important for the cycle consistency objective to work in our few-shot setting. Decoder sharing, in contrast, makes little or even negative influence. This is kinda similar as in multilingual machine translation where sharing the decoder among languages might negatively interfere with the performance~\cite{johnson2017google}.

\begin{table}[t!]
\centering
\resizebox{0.8\columnwidth}{!}{
\begin{tabular}{l|c|c|c|c}

\small\textbf{Text Augmentation}                  & \textbf{\small 1\%}  & \textbf{\small 5\%}   & \textbf{\small 10\% } & \textbf{\small 20\%} \\ \hline
 \small\textbf{None}   & 44.18  & 50.22   & 57.37   & 63.28     \\ \hline
Random                 & 41.30   & 49.62   &  57.71 & 62.79    \\ 
UDA                    & 44.24  & 50.09   &  57.66 & 61.30    \\
Info                   & 45.63  & 52.22   &  58.80 & 63.22    \\ 
+ LM                   & 48.67  & 53.53   & 59.04  & \textbf{64.78}    \\
Reference              & \textbf{55.33}  & \textbf{54.92}   & \textbf{59.11}  & 64.26      \\ \hline\hline
 Random (+RM)          & 42.32  & 50.10   & 58.53  & 63.29    \\
 UDA (+RM)             & 44.32  & 52.22   &  58.80 & 61.27    \\
 Info (+RM)            & 48.63   & 56.66   & 60.80  & 63.52    \\
 + LM (+RM)            & 53.18   & 59.12   & 64.24  & \textbf{65.38}  \\
 Reference (+RM)       & \textbf{62.74}   & \textbf{63.28}   & \textbf{64.93}  & 65.27 \\\hline
\end{tabular}
}
\caption{\label{tb:lmaug}\small Ablation study for text augmentation with varying number of annotations. Experiments are performed on the E2E dataset. \textbf{LM} augmentation outperforms \textbf{Random} by a large margin, and even outperforming augmentation with ground-truth references on some occasions. Representation matching (\textbf{RM}) boosts the overall performance further.  }
\end{table}

\begin{figure*}[h]
\centering
\small
\fbox{
\parbox{0.9\linewidth}{
\textbf{Data}:
\vspace{0.1cm}

[name] Blue Spice [eat type] restaurant [food] Chinese [area] city centre [family friendly] no [near] Rainbow Vegetarian Café  \\
\textbf{Reference}:
at the [city centre], there is a [restaurant] called the [Blue Spice].

\hrule
\vspace{0.1cm}
\textbf{seq2seq/info-aug}: Blue Spice restaurant near Rainbow Vegetarian Café has a \emph{5 star rating. prices start at £30}.
 \\
\textbf{+LM-aug}: located near Rainbow Vegetarian Café is a Chinese theme eatery and restaurant called Blue Spice. It is in the city centre area.
 \\
}}
    \caption{\small Generation examples with different text augmentation techniques. Trained on 5 data annotations (see the above toy training set). The attribute combination of input data is unseen in the 5 annotations. Hallucinated contents are \emph{italitized}.
    }
    \label{fig: example}
\end{figure*}

\begin{figure*}[ht]
  \centering
\includegraphics[width=0.8\textwidth]{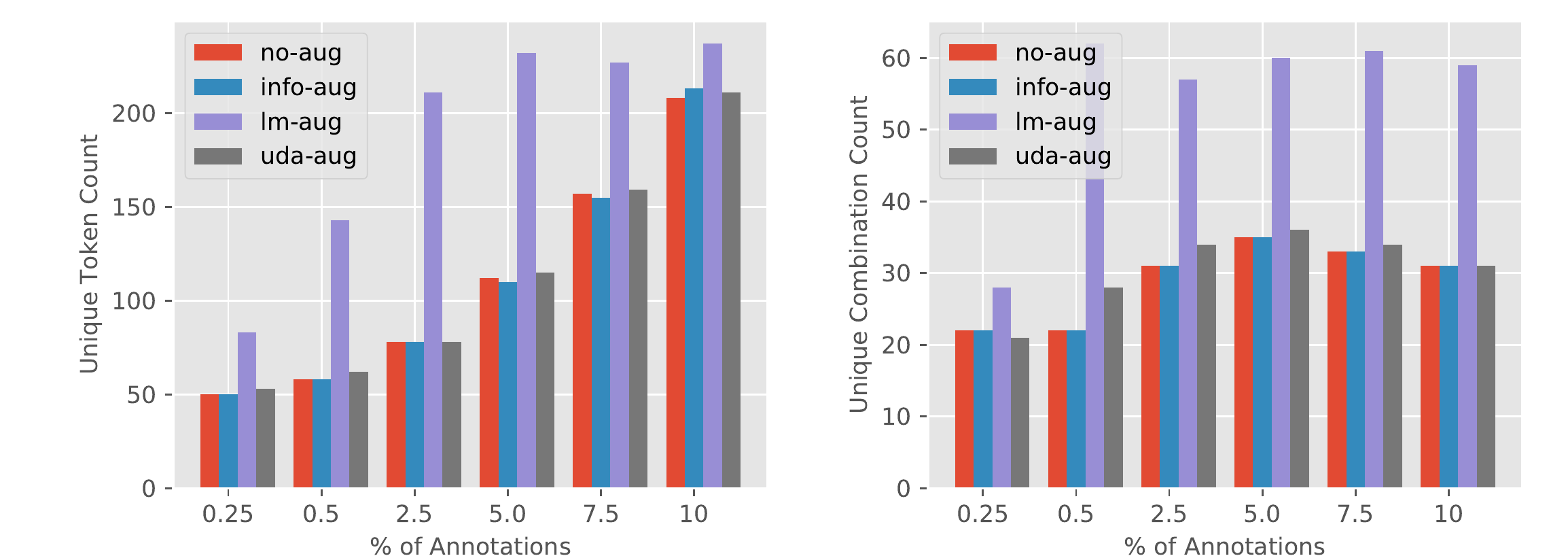}
\caption{\small \label{fig:unique} Additional number of decoded unique tokens (non-copied) and unique combinations of information on the testset with varying number of annotations.}
\label{fig:vocabulary}
\end{figure*}

\begin{table}[t]
\centering
\resizebox{\columnwidth}{!}{%
\begin{tabu}{lccc|ccc}
\tabucline [1pt]{1-7}
\multirow{2}{*}{Model} & \multicolumn{3}{c}{E2E} & \multicolumn{3}{c}{WebNLG} \\
 & Fluency & Miss & Wrong & Fluency & Miss & Wrong \\ \hline
Seq2Seq               & 3.68 & 49 & 63 & 3.95 & 57 & 48 \\
Cycle-only & 4.08 & 46 & 66 & 4.23 & 48 & 44 \\
Finetune GPT-2         & 4.21 & 43 & 57 & 4.10 & 39 & 45 \\
Proposed (LSTM)       & \textbf{4.33} & \textbf{39} & \textbf{44} & \textbf{4.27} & \textbf{31} & \textbf{39} \\
\tabucline [1pt]{1-7}
\end{tabu}%
}
\caption{
\small 
Human Evaluation on the sampled outputs (100 instances) for models with 10\% annotated data. Cycle-only indicates the approach in \citet{qader2019semi}; and Finetuned GPT-2 is refers to \citet{chen2019few}.}
\label{tab: human_evaluation}
    \vspace{-0.3em}
\end{table}

\paragraph{Ablation Study on Text Augmentation}
On top of the four-direction cycle consistency training, we study the effects of text augmentation in Table~\ref{tb:lmaug}. We compare our proposed info + LM augmentation with (1) random augmentation, where a random text from Wikipedia is sampled to the augmented text space, (2) unsupervised data augmentation~\cite{xie2019unsupervised} where text samples are augmented with paraphrases of current annotations and (3) ground-truth augmentation with reference obtained from the left training corpus, which can serve as an upper bound of text augmentation techniques. We test the performance with 1\%, 5\%, 10\% and 20\% annotations to see the effects with varying number of supervisions.

As can be seen, the random augmentation even harms the model performance, suggesting reasonable in-domain text augmentation are necessary for the model improvement. UDA augmentation also makes rather little difference as it simply paraphrases the current available annotations but cannot bring any new information. The information augmentation by slot-value replacement helps improve a bit. When combined with LM, the performance can be further boosted, especially for lower-resource scenarios. The representation matching always helps lift the performance, with gains of up to 10 BLEU points. As expected, the benefit from text augmentation gradually vanishes as more annotations are collected, especially for datasets with relatively simple patterns as E2E.



\paragraph{How text augmentation helps}
Intuitively the GPT-2 augmentation is expected to impose new tokens and combination patterns to the few-shot annotations. To investigate whether this is the case, for the decoded text in the test phase, we count the number of unique tokens (excluding copied data values) and unique information combination patterns (attribute combinations in E2E). The results in Fig.~\ref{fig:unique} show that LM-augmentation indeed greatly enriches the vocabulary space, even doubling the generated unique tokens in low-resource scenarios. The same happens for new combination patterns. In contrast, \emph{info-aug} cannot insert new tokens or combinations at all since all it does is replacing data values based on the same text annotation. UDA can impose new tokens by paraphrasing the annotations, but it hardly helps the model generalize to new combinations of information.
Moreover, when trained on a toy dataset, we observe from the generation outputs that Seq2seq and info-aug produce the wrong outputs and overfit to the information in the 5 training instances. 
With LM augmentation, it adapts to the new combination and connects information correctly. 
Figure~\ref{fig: example} shows a generation example with different text augmentation techniques. We train the systems in a toy setting with only 5 data annotations (Trainset in the Appendix). We pick an input data with an unseen attribute combination to test if models can generalize correctly. Seq2seq and info-aug produce the wrong generation overfit to the information in the 5 training instances. With LM augmentation, it adapts to the new combination and connects information correctly.
\paragraph{Human Evaluation}
We further run a human evaluation on the model outputs to closely check the generation quality. We compared four types of models: the seq2seq baseline, seq2seq plus cycle-consistency as in \citet{qader2019semi}, finetuned GPT-2 as in \citet{chen2019few} and our proposed model. All models are LSTM-based apart from the finetuned GPT-2 one. We sample 100 data instances from the test set and apply all the models to generate corresponding text. The data and generated text are evaluated by $50$ crowdworkers on Prolific\footnote{\url{https://www.prolific.co/}}. For each data-text pair, the annotator is instructed to evaluate (1) if the text is fluent (score 0-5 with 5 being fully fluent), (2) if it misses information contained in the source data and (3) if it includes wrong information. The average fluency scores, count of information miss and wrong information are presented in Table~\ref{tab: human_evaluation}. The scores are generally consistent with the automatic evaluation results, our proposed model outperforms other ones by a large margin, even though cycle-only can access all unpaired text and finetuned GPT-2 is significantly larger than our LSTM-based seq2seq. The generated text are more fluent, yet maintaining the information completeness and correctness to a large extent.

\section{Conclusion}
We study few-shot data-to-text generation with only limited annotated data. We propose text augmentation with slot-value replacement followed by GPT-2 generation. The augmented text, when combined with cycle consistency and representation matching, is shown to help the model to generalize to unseen new tokens and patterns of token combinations. With less than 10\% annotations, it outperforms supervised seq2seq model trained on 100\% annotations and is extensible enough to be combined with pretraining techniques.
For future works, we hope to apply the techniques to annotation scenarios~\cite{hong2019improving,de2018generating,zhuang2017neobility,chang2020unsupervised,wiehr2020safe,shen2020neural,chang2020dart,su2020moviechats,chang2021jointly,chang2021does,crook2018conversational}.

\section*{Acknowledgements}
This research was funded in part by the German Research Foundation (DFG) as part of SFB 248 ``Foundations of Perspicuous Software Systems''. We sincerely thank the anonymous reviewers for their insightful comments that helped us to improve this paper.

\bibliography{anthology,eacl2021}
\bibliographystyle{acl_natbib}

\end{document}